\documentclass[12pt]{article}
\pdfoutput=1
\usepackage{PRIMEarxiv}
\usepackage[T1]{fontenc}
\usepackage[utf8]{inputenc}
\usepackage{listings}
\usepackage{amssymb}
\usepackage{hyperref}

\usepackage{color}
\definecolor{keywordcolor}{rgb}{0.7, 0.1, 0.1}   % red
\definecolor{tacticcolor}{rgb}{0.0, 0.1, 0.6}    % blue
\definecolor{commentcolor}{rgb}{0.4, 0.4, 0.4}   % grey
\definecolor{symbolcolor}{rgb}{0.0, 0.1, 0.6}    % blue
\definecolor{sortcolor}{rgb}{0.1, 0.5, 0.1}      % green
\definecolor{attributecolor}{rgb}{0.7, 0.1, 0.1} % red

\DeclareUnicodeCharacter{2200}{$\forall$}
\DeclareUnicodeCharacter{2203}{$\exists$}
\usepackage{amsthm}
\usepackage{amsfonts}
\usepackage{amsmath}

\usepackage{graphicx}

\newtheorem{thm}{Theorem}

\title{GFLean: An Autoformalisation Framework for Lean via GF
}

\author{
    Shashank Pathak \\
  Department of Computer Science \\
  The University of Manchester \\
  Manchester, UK \\
  \texttt{shashank.pathak@manchester.ac.uk} \\
}

\begin{document}
\maketitle

\begin{abstract}
We present an autoformalisation framework for the Lean theorem prover, called GFLean.
GFLean uses a high-level grammar writing tool called Grammatical Framework (GF) for parsing and linearisation.
GFLean is implemented in Haskell.
We explain the functionalities of GFLean, its inner working and discuss its limitations.
We also discuss how we can use neural network based translation programs and rule based translation programs together complimenting each other to build robust autoformalisation frameworks.
\end{abstract}

% keywords can be removed

\section{Introduction}
Formalisation refers to the task of converting a text from a natural language to a formal language.
Formalisation of mathematical text is beneficial for two reasons.
Firstly, formalisation is done in an underlying logical system and thus is accompanied by proof checking.
Secondly, formalised mathematical text can be stored and manipulated by computers.
Mechanizing the process of formalisation is called \textit{autoformalisation}.
In this article, we present our ongoing work on creating an autoformalisation framework, called GFLean.
GFLean uses a high-level grammar writing tool called Grammatical Framework (GF) \cite{ranta2011grammatical}, and converts simple statements from the \textit{language of mathematics} to input for the Lean theorem prover \cite{moura2021lean}.

By the language of mathematics, we mean the text found in pure mathematics textbooks and research articles \cite{ganesalingam2013language}.
In his PhD thesis, M. Ganesalingam \cite{ganesalingam2013language} distinguishes between two classes of sentences found in the language of mathematics, called the \textit{formal mode} and the \textit{informal mode}.
The sentences in the formal mode have a strict mathematical content and can be formalised, whereas those in the informal mode have the purpose of guiding the reader's intuition and giving a commentary on the text.
For example, the sentence ``The Lagrange theorem is important and will be useful later'' is in the informal mode.
For GFLean, we only work with sentences in the formal mode.

\noindent GFLean is implemented in Haskell and uses GF for parsing the input and linearizing the output.
An example of a GFLean input is
\begin{verbatim}
    Ex. Assume x is a rational number. Assume x is equal to 2 + 2 * 2. Then x is greater
    than 3.
\end{verbatim}
The corresponding output is
\begin{lstlisting}
    example (x : ℚ) (h39 : x = (2 + (2 * 2))) : x > 3 := sorry
\end{lstlisting}
Currently, GFLean only formalises statements but not proofs.
The code for GFLean and some working examples can be found on the GitHub repository of the project \cite{gflean}.

We call the input language accepted by GFLean \textbf{Simplified ForTheL}.
Simplified ForTheL is based on the controlled natural language ForTheL \cite{vershinin2000forthel}.
The ForTheL syntax has already been implemented in GF \cite{schaefer2020prototyping} and our implementation of the Simplified ForTheL syntax in GF is based upon the ForTheL syntax implementation.
However, the semantics of ForTheL expressions \cite{vershinin2000forthel} and Simplified ForTheL expressions differ. 
ForTheL expressions are converted to first-order formulas, whereas Simplified ForTheL expressions are converted to Lean expressions, which are type-theoretic in nature.
Thus, we give a type-theoretic semantics to Simplified ForTheL expressions.
Specifically, our contributions are the following.
\begin{enumerate}
    \item Implementing algorithms in Haskell to translate Simplified ForTheL expressions to Lean expressions via manipulating \textbf{abstract syntax trees} (ASTs).
    \item Implementing a grammar for Lean expressions in GF, so that the ASTs obtained from the previous step can be linearized as Lean expressions. 
\end{enumerate}

\noindent To test how GFLean performs on statements from a standard textbook, we used GFLean to formalise statements from Chapter 3 of the introduction to proofs textbook \textit{Mathematical Proofs} by G. Chartrand, A. D. Polimeni, and P. Zhang \cite{chartrand2017mathematical}.
Out of the 62 statements contained in the chapter, GFLean can parse and formalise 42 statements with a minor rephrasing of the input.
The 42 statements, the corresponding GFLean inputs, and the GFLean outputs are given in Appendix \ref{examples}.
To parse the remaining 20 statements, the lexicon needs to be expanded and GFLean needs to have support for more linguistic phenomena like post-fix quantification of symbols, donkey anaphora, etc.

This article is structured in the following manner.
Section \ref{relatedWorks} gives a brief overview of the work done in autoformalisation of mathematical text.
It also contains a brief introduction to the Lean theorem prover and a detailed explanation of how a GF program works.
Section \ref{gflean} gives a thorough description of the workings of GFLean.
Section \ref{limitations} outlines the limitations of GFLean.
Section \ref{discussion} discusses how rule-based systems and neural translation systems can be used together to create robust autoformalisation tools.
Section \ref{conclusion} concludes the article and mentions the directions in which we plan to extend GFLean.
Appendix \ref{examples} contains the input and output for the 42 statements which GFLean can formalise from the textbook mentioned above.
Appendix \ref{grammar} contains a formal grammar for Simplified ForTheL.

\section{Related Work and Preliminaries} \label{relatedWorks}
Significant work has been done in writing and checking proofs written in a language that looks like natural language.
C. Zinn \cite{zinn2004understanding} used an extension of discourse representation theory \cite{kamp2010discourse} called proof representation structures to represent mathematical discourse.
The corresponding system, called Vip, was able to process two full textbook proofs, but was a proof of concept.
The System for Automated Deduction (SAD) \cite{verchinine2007system} was developed to write and check proofs written in the controlled natural language (CNL) ForTheL \cite{vershinin2000forthel}.
SAD converts ForTheL expressions to first-order formulas and passes them to first-order automated theorem provers for checking.
An ongoing project with the same objectives of writing and checking proofs in a CNL is Naproche \cite{de2021isabelle}.
The CNL used as an input for Naproche is also based upon ForTheL, and comes with a \LaTeX{} dialect.
Thus, the mathematical documents written for Naproche also get typeset as \LaTeX{} documents.
A number of results, like group theory up to the Sylow theorems, initial chapters from Walter Rudin’s Analysis, and set theory up to Silver’s theorem in cardinal arithmetic have been formalised and verified in Naproche.
 
The language of mathematics from a linguistic point of view was studied by Ganesalingam in his PhD thesis \cite{ganesalingam2013language}.
Ganesalingam mentions the linguistic features found in the language in detail and gives a blueprint of formalisation mechanisms using an extended version of DRT.
As far as we know, the work has not been implemented as a practical tool.

Two projects which use Grammatical Framework (GF) for autoformalisation deserve a mention here.
The first, called the MathNat project \cite{humayoun2010mathnat}, consisted of developing a CNL for writing statements and proofs, and developing a system-independent abstract mathematical language.
In MathNat, GF was used to parse the CNL expressions.
The second is the Grammatical Logical Inference Framework (GLIF) \cite{schaefer2020glif}, which can be used to do inference on statements written in a CNL.
Here too, GF is used for parsing, but other tools are used for semantic construction, logical processing, and inference.

Apart from rule-based systems, neural networks have also been used for autoformalisation.
Neural translation methods have been used to translate {\LaTeX} strings to Mizar input \cite{wang2018first},\cite{wang2020exploration}.
In recent work, Large language models (LLMs) were used to formalise mathematical problems in to Isabelle/HOL input \cite{wu2022autoformalization} by providing a few examples and asking the LLM to formalise the statement.
Similarly, the LLM Codex was used, alongwith input-dependent prompt selection, to formalise 120 natural language statements to Lean input  with a 65\% accuracy \cite{gadgil2022towards}.  
Generating parallel corpora for training models is an expensive task, although recently LLMs have been used for that as well.
A corpus of 332K formal-informal statement pairs has been produced \cite{jiang2023multilingual} by informalising statements using GPT-4 \cite{achiam2023gpt} from the Archive of Formal Proofs \cite{Archive_of_formal_proofs_1970}, which is a collection of proof libraries for Isabelle, and the Lean 4 library {\fontfamily{cmss}\selectfont mathlib4} \cite{mathlib4}.

\subsection{Lean}

The \textbf{Lean 4} theorem prover (or Lean) is an interactive theorem prover and a full-fledged programming language \cite{moura2021lean}.
Lean and its predecessors have been used for large-scale formalisation projects like the \textit{Sphere Eversion Project} \cite{van2023formalising} and \textit{Perfectoid Spaces} \cite{buzzard2020formalising}.
Lean also has a rapidly evolving monolithic mathematical library called {\fontfamily{cmss}\selectfont mathlib4}.
As of March 2024, {\fontfamily{cmss}\selectfont mathlib4} has about 140k theorem statements and 76k definitions \cite{Mathlibstatistics}.
For GFLean, we chose the target language as Lean because Lean can be used to formalise research-level mathematics, and a large body of mathematics has already been formalised in it.

\subsection{Grammatical Framework (GF)}
\textbf{Grammatical Framework} (GF) is a special-purpose programming language designed to write multilingual grammars.
Each GF program is called a GF grammar and is made up of a single \textbf{Abstract Syntax} and at least one \textbf{Concrete Syntax}.
For translation, the abstract syntax acts as a bridge between the various concrete syntaxes.
The abstract syntax encodes everything that needs to be preserved during translation.
The concrete syntaxes encode language specific peculiarities, for example number or gender agreements.

Building translation systems via GF uses one of the two following methods.
\begin{enumerate}
    \item \textbf{Using only GF.}
    In this method, a single abstract syntax is defined, and as many concrete syntaxes are defined as there are languages.
    The GF programmer has to define the abstract syntax and concrete syntaxes in a way such that the syntax for all the languages can be realized as the concrete syntaxes corresponding to the same abstract syntax.

    \item \textbf{Embedding GF grammars in a host program.} One usually adopts this method if in order to carry out a faithful language translation, defining a common abstract syntax is difficult, and one needs to perform some abstract syntax tree (AST) transformations.
    In this method, a GF grammar is used to parse the input or linearize the output, but an external host program is used to perform the AST transformations.
    A. Ranta \cite{ranta2011translating}, who has played a leading role in the development of GF, mentions that GF lacks the program constructs and libraries needed for non-compositional translations, such as list processing, state management, and so on, but these translations can be done via embedding a GF grammar in a host program.
    GF grammars can be embedded in Haskell, Python and JavaScript programs \cite{ranta2011grammatical}.
    Simple but useful interfaces for formalising mathematics employing this method have already been built \cite{ranta2011translating}.
    Since for GFLean we had to implement some non-compositional translations, we chose this method.
    Specifically, for GFLean we used GF to parse the input and linearize the output, and we used Haskell to do the AST transformations.
    
\end{enumerate}

\noindent GF is a high-level grammar writing tool.
The user just needs to write the grammar rules, and gets the lexer, parser, and type-checker for free \cite{ranta2011grammatical}.
The user can utilize records and tables to define concrete syntaxes such that the various grammatical agreement rules hold, such as the grammatical number agreement between the verb phrase and the noun phrase, or gender agreement between a noun and a pronoun.
The division of functionalities between the abstract and concrete syntax makes grammar writing modular.
Along with that, the user can import natural language grammars as software libraries from the GF Resource Grammar Library (RGL).
As of 2019, RGL has implementations of 35 natural languages \cite{ranta2020abstract}.
For GFLean, we chose not to import RGL because the natural language lexicon in GFLean is small and we are not focusing on any language except English.
Next, we explain the abstract syntax, the concrete syntax in detail via an example.

\subsubsection{Abstract Syntax} In a GF grammar, the abstract syntax defines the linguistic categories and the possible parse trees.
The following example defines an abstract syntax, called \texttt{Demo}, for a toy grammar for the language of mathematics.

\begin{verbatim}
    abstract Demo = {
    cat
        Prop; Var;
    fun
        Nzero, Greater1 : Var -> Prop;
        Imp : Prop -> Prop -> Prop;
        ForAll, Exists : Var -> Prop -> Prop;
        Var1, Var2 : Var;
    }
\end{verbatim}

\noindent For \texttt{Demo}, the lingusitic categories are \texttt{Prop} and \texttt{Var}, which stand for propositions and variables respectively.
Under \texttt{fun}, we define the type of each of the constructors.
For example, \texttt{Nzero} has the type \texttt{Var -> Prop} which means \texttt{Nzero} can be applied to a term of type \texttt{Var} to produce a term of type \texttt{Prop}.
The term \texttt{Var1} has type \texttt{Var}.
Thus, \texttt{Nzero Var1} has type \texttt{Prop}.

\subsubsection{Concrete Syntax} A concrete syntax defines how the trees defined via the abstract syntax are linearized as strings.
Consider the following concrete syntax called \texttt{DemoEng}, which defines how the abstract syntax trees (ASTs) defined by \texttt{Demo} are linearized as English sentences.

\begin{verbatim}
    concrete DemoEng of Demo = {
    lincat
        Prop, Var = Str;
    lin
        Nzero var = var ++ "is nonzero";
        Greater1 var = var ++ "is greater than 1";
        Imp prop1 prop2 = "If" ++ prop1 ++ ", then"
                        ++ prop2;
        ForAll var prop = "For each" ++ var ++", " ++ prop;
        Exists var prop = "There exists an" ++ var
                        ++ "such that" ++ prop;
        Var1 = "x1";
        Var2 = "x2";
    }
\end{verbatim}

\noindent Here, \texttt{Str} is the built-in string GF datatype, and \texttt{++} is the string concatenation operator. 
The following is another concrete syntax called \texttt{DemoMath} which linearizes the ASTs from \texttt{Demo} into symbolic mathematics.

\begin{verbatim}
    concrete DemoMath of Demo = {
    lincat
        Prop, Var = Str;
    lin
        Nzero var = "~ ("++ var ++ " = 0 )";
        Greater1 var = var ++ " > 1";
        Imp prop1 prop2 = "(" ++ prop1 ++ "→" ++ prop2 ++ ")";
        ForAll var prop = "(∀" ++ var ++", " ++ prop ++ ")";
        Exists var prop ="(∃" ++ var ++ "," ++ prop ++ ")";
        Var1 = "x1";
        Var2 = "x2";
    }
\end{verbatim}      

\begin{figure}[h]
    \centering
    \includegraphics[scale=0.4]{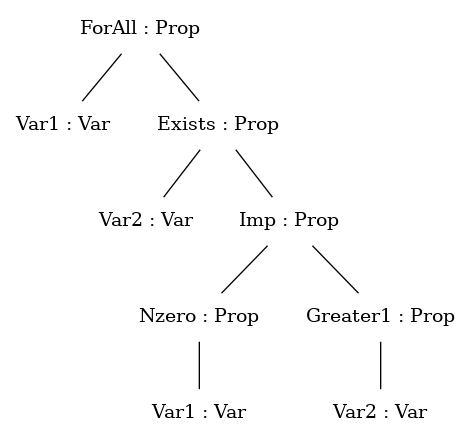}
    \caption{The AST produced by GF for a sentence from \texttt{DemoEng}.}
    \label{fig:ASTDemo}
\end{figure}

\noindent To translate a string $s$ from a language $L_1$ to $L_2$, GF uses the concrete syntax for $L_1$ to parse $s$, and find a corresponding AST.
Then, GF linearizes the AST as a string of $L_2$.
For example, for the \texttt{Demo}, \texttt{DemoEng}, and the \texttt{DemoMath} grammar, parsing the sentence 
\begin{verbatim}
    For each x1 , There exists an x2 such that If x1 is nonzero , then x2 is greater
    than 1 
\end{verbatim}
produces the AST as shown in Figure \ref{fig:ASTDemo}.
Then, GF linearizes the produced AST as
\begin{verbatim}
    (∀ x1 , (∃ x2 , ( ~ ( x1 = 0 ) → x2 > 1 ) ) ).
\end{verbatim}

\section{GFLean} \label{gflean}

\begin{figure}[h]
    \centering
    \includegraphics[scale = 0.4]{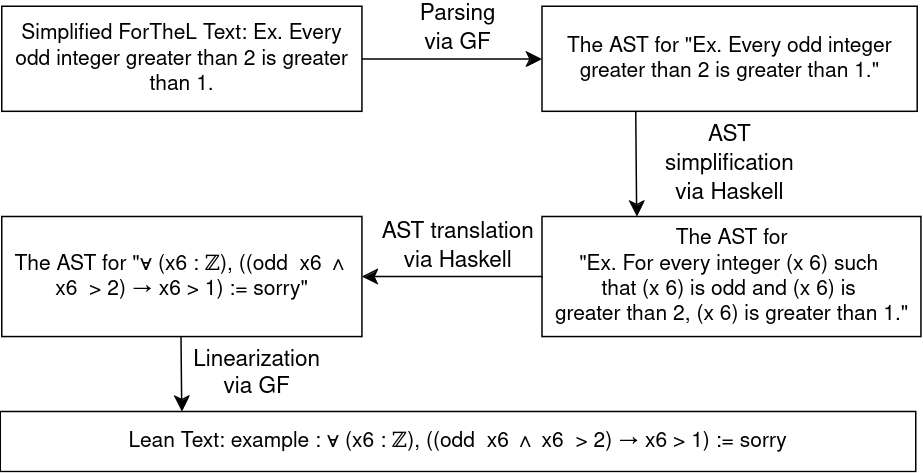}
    \caption{The GFLean processing pipeline}
    \label{fig:GFLean}
\end{figure}

GFLean is implemented Haskell and contains two separate embedded GF grammars.
The first grammar defines the input for GFLean.
We call the input language for GFLean Simplified ForTheL.
The second grammar is used to linearize the tranlsated abstract syntax trees (ASTs) to Lean expressions.
GFLean performs the following steps in the given order to convert a Simplified ForTheL expression to a corresponding Lean expression:
\begin{enumerate}
    \item \textbf{Parsing}: GFLean parses the Simplified ForTheL expression using a GF grammar and produces the expression's AST.
    \item \textbf{Simplification}: A series of tree transformations, which we call simplification, happen on the AST.
    The simplification is implemented in Haskell.
    \item \textbf{Translation}: The AST for the Simplified ForTheL expression obtained from the previous step is converted into an AST for the corresponding Lean expression.
    This step is implemented in Haskell as well.
    \item \textbf{Linearization}: The AST for the Lean expression is linearized as the Lean expression using GF.
\end{enumerate}
The translation pipeline comprising the four steps above is shown in Figure \ref{fig:GFLean}.

The adopted method employed in GFLean of simplifying the ASTs and then translating them into target expressions is broadly taken from the System for Automated Deduction (SAD) \cite{verchinine2007system}.
The methodological differences between SAD and GFLean are the following:
\begin{enumerate}
    \item SAD accepts ForTheL, whereas GFLean only accepts Simplified ForTheL.
    \item GFLean uses GF for parsing and linearization, whereas in SAD all the steps are fully implemented in Haskell.
\end{enumerate}

\subsection{Simplified ForTheL}

Simplified ForTheL is a simplified version of the controlled natural language ForTheL \cite{vershinin2000forthel}.
ForTheL is the input language used by SAD \cite{verchinine2007system}, and the input language for Naproche is based upon ForTheL \cite{de2021isabelle}. 
Our GF implementation of the Simplified ForTheL syntax is based upon a GF implementation of the ForTheL syntax used in the Grammatical Logical Framework ({\fontfamily{cmss}\selectfont GLF}) \cite{schaefer2020prototyping}.

\subsubsection{Differences between ForTheL and Simplified ForTheL}\label{diff}
Simplified ForTheL is a simplified version of ForTheL in the following sense:
\begin{enumerate}
    \item \textbf{Left adjectives.} In ForTheL, multiple left-adjectives can be added in front of an entity, whereas in Simplified ForTheL, only one left-adjective is allowed.
    For example, in ForTheL one can write \texttt{x is an odd prime integer}, but in Simplified ForTheL one has to write  \texttt{x is an integer, x is odd and x is prime}.
    
    \vspace{1mm}
    \item \textbf{Conjunction of predicates.} In ForTheL, a conjunction of predicates is allowed but in Simplified ForTheL, only a single predicate in a sentence is allowed.
    For example, in ForTheL one can write \texttt{x is odd and greater than 4}, but in Simplified ForTheL one has two write \texttt{x is odd and x is greater than 4}.
    
    \item \textbf{Conjunction of terms.} In ForTheL, a clause can have a conjunction of terms as the subject but in Simplified ForTheL, a clause can just have a single term as the subject.
    Thus, in ForTheL one can write \texttt{x and y are odd} but in Simplified ForTheL one has to write \texttt{x is odd and y is odd}.

    \item \textbf{Macro-grammar.} By macro-grammar, we mean how the sentences are organised to form a text on the whole.
    The macro-grammar for ForTheL is geared towards SAD but the macro-grammar for Simplified ForTheL is geared towards Lean.

    \item \textbf{Dynamicity.} The lexicon for ForTheL is dynamic in the sense that the user can add to the lexicon during runtime by using patterns \cite{vershinin2000forthel}.
    Currently, GF grammars are static and can not be changed during runtime.
    Thus, it is not possible expand the Simplified ForTheL lexicon during runtime.
\end{enumerate}

\subsection{GFLean Translation Examples}
We will see a few input-output examples produced by GFLean.
The input is processed in a series of steps each of which is shown when GFLean is run.
For brevity, here we just show the input, and the final output.
The examples demonstrate in detail the range of natural language expressions which GFLean can process.

The input
\begin{verbatim}
    Ex. Assume x is a rational number equal to 2 * 2. Then x is greater than 3.
\end{verbatim}
produces the output
\begin{lstlisting}
    example (x : ℚ) (h33 : x = (2 * 2)) : x > 3 := sorry
\end{lstlisting}
Thus, GFLean can process expressions containing basic arithmetic operators.
The input
\begin{verbatim}
    Ex. Assume x is a real number less than 0. Then no nonnegative integer a such that a
    is positive is not greater than x.
\end{verbatim}
produces the output
\begin{lstlisting}
    example (x : ℝ) (h67 : x < 0) : ∀ (a : ℤ), ((nneg a ∧ pos a) → (¬ (¬ a > x))) := sorry
\end{lstlisting}
Thus, GFLean can correctly model how the natural language quantifiers (in this case \texttt{no}) and negation occurring inside a sentence (\texttt{not}) interact.
The input
\begin{verbatim}
    Ex. Assume x is an even integer greater than 32. Then x is greater than every integer
    less than 32.
\end{verbatim}
produces the output
\begin{lstlisting}
    example (x : ℤ) (h70 : even x) (h56 : x > 32) : ∀ (x34 : ℤ), (x34 < 32 → x > x34) := sorry
\end{lstlisting}
Thus, we can modify common nouns like \texttt{integers} with adjectives to the left (in this case \texttt{even}) and adjectival phrases to the right (in this case \texttt{greater than 32}) in the input.
Also, we can have quantifiers in the predicate (in this case \texttt{every} in \texttt{every integer less than 32}).

\subsection{The Workings of GFLean}
In this section, we explain the four steps mentioned before in detail.

\subsubsection{Parsing Simplified ForTheL expressions}
\begin{figure}[h]
    \centering
    \includegraphics[width=\textwidth]{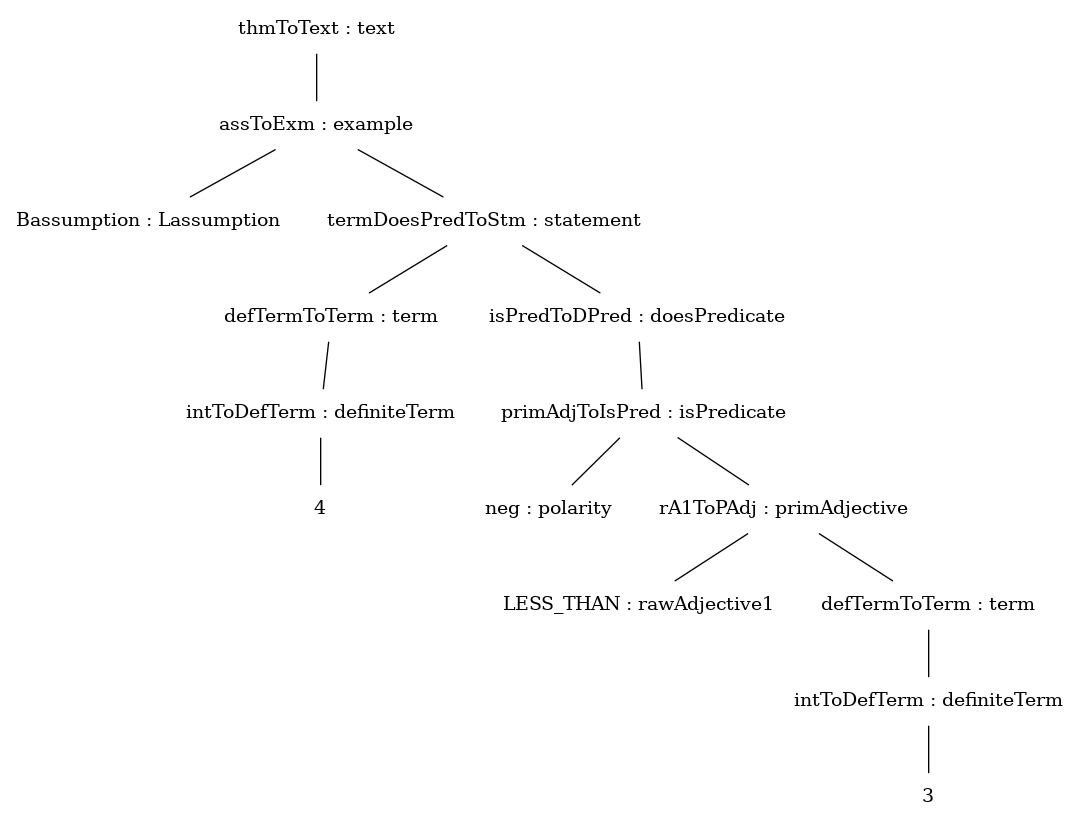}
    \caption{The AST produced by GF after parsing \texttt{Ex.\;4 is not less than 3.}}
    \label{fig:AST}
\end{figure}

The parsing of a Simplified ForTheL expression is done via GF for which we had to implement the syntax of Simplified ForTheL as a GF grammar.
GF parses the expression and produces an abstract syntax tree (AST) which is passed on to the next step.
For example, after parsing the Simplified ForTheL expression
\begin{verbatim}
    Ex. 4 is not less than 3.
\end{verbatim}
the AST shown in Figure \ref{fig:AST} is passed on to the next step.

Currently, the lexicon contains 8 items like \texttt{REAL\_NUMBER, INTEGER, EXP, SUM, MINUS,} etc. which, after combining with one or multiple terms, behave like a noun phrase, and 11 items like \texttt{LESS\_TE, GREATER\_THAN, POSITIVE,} etc. which, after combining with one or multiple terms, behave like an adjectival phrase.
For both the Simplified ForTheL grammar and the Lean grammar, we extract the lexicon from the same GF file, which makes grammar extension more modular.

We endow the arithmetic operators with a precedence hierarchy by using records in the concrete syntax of the Simplified ForTheL grammar.
As a result, the Simplified ForTheL expression \texttt{2 + 2 * 2} has the same AST as the expression \texttt{2 + (2 * 2)}.
To override the precedence hierarchy, the user needs to use \texttt{(} and \texttt{)}.
For example, in this case the user needs to write \texttt{(2 + 2) * 2}.

\subsubsection{AST simplification}
This step works on the level of ASTs, i.e. it makes certain changes the to ASTs produced by the previous step.
Because of space limitations, instead of showing how the simplification changes the ASTs, we will show how the linearizations of the ASTs changes during simplification.
This step is implemented in Haskell, and is made up of many intermediate sub-steps.
We explain the sub-step in detail now.

\noindent The first sub-step is giving an entity a name if it is unnamed.
This sub-step is present in SAD \cite{vershinin2000forthel} as well.
The name of an unnamed entity in the Simplified ForTheL grammar is represented by a metavariable.
We replace the metavariable with an actual name.
The names introduced are new and differ from the names already used.
For example, after this sub-step, the AST for
\begin{verbatim}
    Ex. Assume x is an integer. Assume x is greater than 2. Then no odd integer less than
    1 is greater than x.
\end{verbatim}
becomes the AST for
\begin{verbatim}
    Ex . Assume x is a integer (x 6). Assume x is greater than 2. Then no odd integer
    (x 35) less than 1 is greater than x.
\end{verbatim}
We call the next sub-step variable unification.
This sub-step is needed for correct translation because sometimes the introduced names need to match the variables names already present in the sentence.
For example, after this sub-step, the AST for
\begin{verbatim}
    Assume x is a rational number (x 6).
\end{verbatim}
becomes the AST for
\begin{verbatim}
    Assume x is a rational number x.
\end{verbatim}
 
\noindent In the next sub-step, we change the ASTs such that the corresponding sentences are in a certain form without changing the meaning.
Specifically, we convert \textit{in-situ quantification} to \textit{ex-situ quantification}.
Consequently, the AST for
\begin{verbatim}
    Every integer x greater than 1 is greater than 2.
\end{verbatim}
becomes the AST for
\begin{verbatim}
    For every integer x greater than 1, x is greater than 2.
\end{verbatim}
Then, we change the structure of the entities so that both the left-adjectives and the adjectival phrases present on the right of a noun are written just as a statements modifying the noun from the right.
For example, the AST for 
\begin{verbatim}
    odd integer x greater than 1
\end{verbatim}
becomes the AST for
\begin{verbatim}
    integer x such that x is odd and x is greater than 1
\end{verbatim}

\noindent After the executing the AST simplification step fully, the AST for
\begin{verbatim}
    Ex. Assume x is an integer. Assume x is greater than 2. Then no odd integer less than
    1 is greater than x.
\end{verbatim}
becomes the AST for
\begin{verbatim}
    Ex . Assume x is an integer x. Assume x is greater than 2. Then for no integer (x 35)
    such that (x 35) is odd and (x 35) is less than 1, (x 35) is greater than x.
\end{verbatim}

\subsubsection{AST transformation}
Similar to AST simplification, in this step too, we make changes to the AST.
But because of space limitations, we show the changes in the corresponding linearized strings instead of the ASTs themselves.
In this step, we take the ASTs obtained after performing the simplification and construct new ASTs corresponding to the correct Lean translation.
We first translate the lexicon by translating linguistic categories to unary or binary functions and predicates.
Specifically,
\begin{enumerate}
    \item We call the lexical item a \textit{rawNoun0} which behaves like a noun phrase.
    A \textit{rawNoun0} is converted into a Lean type.
    For example, the lexical item \texttt{INTEGER}, which is a \textit{rawNoun0}, gets converted to the type \lstinline{Int}.
    
    \item We call the lexical item a \textit{rawNoun2} which after taking two terms behaves like a noun phrase. 
    A \textit{rawNoun2} is converted into a binary function.
    For example, consider the lexical item \texttt{EXP}, which is a \textit{rawNoun2} and corresponds to the exponent function.
    After taking two terms, \texttt{EXP} behaves like a noun phrase (e.g. \texttt{EXP 2 3}, which can be used in place of a noun as it is equal to 8).
    \texttt{EXP} gets converted to the Lean exponent function which takes two arguments.

    \item We call the lexical item a \textit{rawAdjective0} which behaves like an adjectival phrase.
    A \textit{rawAdjective0} is converted into a unary relation.
    For example, the lexical item \texttt{POSITIVE}, which is a \textit{rawAdjective0}, gets converted to the unary relation $P$, where $P x$ stands for $x$ is positive.

    \item We call the lexical item a \textit{rawAdjective1} which after taking a term behaves like an adjectival phrase.
    A \textit{rawAdjective1} gets converted into a binary relation.
    The lexical item \verb|LESS_THAN|, which is a \textit{rawAdjective1}, gets converted to the binary relation $L$, where $L x y$ signifies $x$ is less than $y$.
\end{enumerate}
For sentences, the ASTs obtained after simplification get converted to ASTs for Lean expressions in a meaning preserving way.
For example, the input
\begin{verbatim}
    Ex. Assume x is an odd integer greater than 3. Then x is greater than 2.
\end{verbatim}
after being parsed and going through simplification, produces the AST for the expression
\begin{verbatim}
    Ex. Assume x is an integer x. Assume x is odd. Assume x is greater than 3. Then x is
    greater than 2.
\end{verbatim}
which in turn gets translated to the AST for the Lean expression
\begin{lstlisting}
    example (x : ℤ) (h40 : odd x) (h27 : x > 3) : x > 2 := sorry
\end{lstlisting}
after the AST translation step.
The natural language quantifiers and the logical connectives contained in the input are also translated in a meaning-preserving manner.
For example, the input
\begin{verbatim}
    Ex. Assume x is an odd integer greater than 3. Then no even integer greater than x is
    less than every negative integer.
\end{verbatim}
after being parsed and going through simplification, produces the AST for the expression
\begin{verbatim}
    Ex. Assume x is an integer x. Assume x is odd. Assume x is greater than 3. Then for no
    integer (x 32) such that (x 32) is even and (x 32) is greater than x, for every
    integer (x 53) such that (x 53) is negative, (x 32) is less than (x 53).
\end{verbatim}
which in turn get translated to the AST for the Lean expression
\begin{lstlisting}
    example (x : ℤ) (h111 : odd x) (h98 : x > 3) : ∀ (x32 : ℤ),
    ((even x32 ∧ x32 > x) → (¬ ∀ (x53 : ℤ), (neg x53 → x32 < x53))) := sorry
\end{lstlisting}
after the AST translation step.

\subsubsection{Linearizing as Lean expressions} The last step of linearizing the ASTs to Lean expression is done by GF.
For this step, we had to write a GF grammar for Lean expressions.

Along with the four steps mentioned in this section, we do a minor pre-processing to the input and a minor post-processing to the output.
The pre-processing steps include converting all text to lowercase, and the post-processing steps include deleting extra whitespaces and giving each hypothesis variable a unique name.

\section{Limitations} \label{limitations}
GFLean still is a rudimentary program, has a tiny lexicon and accepts a small fragment of the language of mathematics.
Regarding the Simplified ForTheL concrete syntax, we use variations for singular and plural forms of nouns and verbs.
As a result, in the Simplified ForTheL concrete syntax there is no difference between \texttt{is} and \texttt{are}, \texttt{a} and \texttt{an}, \texttt{integer} and \texttt{integers}, etc.
Thus, GFLean can accept ungrammatical sentences like \texttt{Assume x are an odd integers.}

As mentioned in Section \ref{diff}, Simplified ForTheL lacks certain linguistics constructs such as conjunction of predicates, conjunction of terms and multiple left-adjectives.
These constructs are abundant in the language of mathematics, and are present in ForTheL as well.
Thus, Simplified ForTheL is not an adequate controlled natural language for the language of mathematics.

Another limitation concerns how much the user can expand the lexicon themselves.
The language of mathematics is dynamic in the sense that definitions and notations introduce new grammar rules and words to the lexicon \cite{ganesalingam2013language}.
Thus, any functioning grammar for the language of mathematics should be extensible via definitions.
A limitation of GF is that the grammars cannot be extended during run-time.
As a consequence, the user cannot write new definitions and convert them to Lean expressions using GFLean.
All the definitions need to be hard-wired in the grammar.

\section{Discussion} \label{discussion}
Continuing the discussion about dynamicity from Section \ref{limitations}, GFLean nowhere communicates with Lean.
One possible way to attain dynamicity would be to reimplement GFLean in Lean itself.
Lean is the target language for GFLean, and is itself a full-fledged programming language \cite{moura2021lean}.
The metaprogramming features of Lean allow us to access the environment and use the definitions and theorems in other Lean programs.
Thus, dynamicity should not be hard to achieve once GFLean has been implemented in Lean, although how hard it is to reimplement GFLean in Lean is an open question.

On another note, rule-based translation systems and neural translation systems can be used together to build more robust autoformalisation programs.
Building rule-based translation systems amounts to manually designing numerous translation rules.
Neural network based translation systems do not have this problem, but are sometimes erroneous and thus domain experts are needed to filter out the incorrect translations.
Rule-based systems designed for linearizing formal expression to natural language text can be used to help the user filter out the wrong translations without putting a domain expert in the loop.
For example, given a natural language statement $s$, let $s'$ be the formalisation of $s$ produced by a neural translation system.
To check whether $s'$ is indeed a correct formalisation of $s$, we need a domain expert.
But if a rule-based system, which correctly linearizes the formal expressions into natural language statements, is available, the rule-based system can be used to convert $s'$ to a natural language statement $s''$.
The user can then themselves filter out a wrong translation by checking if $s$ and $s''$ mean the same thing.
This eliminates the need of a domain expert and results in a more robust autoformalisation program.

\section{Conclusion and Further Work} \label{conclusion}
In this article, we have presented our ongoing effort to construct a framework for autoformalisation, called GFLean.
GFLean converts simple mathematical statements to expressions for the Lean theorem prover.
We use a high-level grammar writing tool called GF for parsing and linearisation.
Using GF allows us to just design the grammar, and we get the tokenizer and parser for free.
For the intermediate steps, we use Haskell to perform abstract syntax tree manipulations.
GFLean is still a program under development and the grammar for the input is very basic.
GFLean can handle simple natural language quantifiers, logical operations, adjectival modifications and quantifiers occurring in the predicate in the input, but does not have support for sentences with a conjunction of predicates, or a conjunction of terms, or more than one adjective modifying a noun.
In terms of how GFLean performs on examples from a textbook, it can parse and formalise 42 out of the 62 statements from Chapter 3 of the textbook \textit{Mathematical Proofs} by G. Chartrand, A. D. Polimeni, and P. Zhang \cite{chartrand2017mathematical} with minor rephrasing.
The preliminary work outlined here supports our working assumption that GF is a useful tool to build modular and potentially scalable rule-based autoformalisation programs.

We plan to extend GFLean in the following directions.
\begin{enumerate}   
   \item We want to make both the concrete syntaxes better by using records, tables and parameters. 
    By using these high-level constructs, we can model the agreements found in English.

    \item ForTheL has some of the linguistic constructs  that Simplified ForTheL lacks.
    We want to extend GFLean from Simplified ForTheL to ForTheL.
    
    \item We want to expand the lexicon.
\end{enumerate}

\section*{Acknowledgements}
The work is supported by the Faculty of Science and Engineering, The University of Manchester.
The author is thankful to Dr. Ian Pratt-Hartmann for his expertise, insights, helping with the plan of the project and proof-reading the manuscript.
The author is also thankful to Dr. Inari Listenmaa for help with GF-related queries, and to Jan Frederik Schaeffer for stimulating discussions and help with the GF implementation of Simplified ForTheL.

\bibliographystyle{splncs04}
\bibliography{references}

\newpage
\appendix

\section{Formalisation of statements from a textbook via GFLean} \label{examples}
GFLean can formalise 42 out of 62 statements from Chapter 3 of the textbook \textit{Mathematical Proofs} by G. Chartrand, A. D. Polimeni, and P. Zhang \cite{chartrand2017mathematical}.
Next, we show how each of the statement can be formalised using GFLean.
We present them in the following manner:

\textbf{Theorem Number} (Result number in the book.)
    Statement from the book.
\begin{verbatim}
    A corresponding input for GFLean.
\end{verbatim}
\begin{lstlisting}
    The corresponding GFLean output.
\end{lstlisting}

The following are the formalisations. 

%Theorem 1
\begin{thm}[Result 3.1]
    Let $x \in \textbf{R}$.
    If $x < 0$, then $x^2 + 1 > 0$.
\end{thm}
\begin{verbatim}
    Ex. Assume x is a real number. Assume x is less than 0. Then x ^ 2 + 1 is greater
    than 0.
\end{verbatim}
\begin{lstlisting}
    example (x : ℝ) (h39 : x < 0) : ((x ^ 2) + 1) > 0 := sorry
\end{lstlisting}

\hrule
%Theorem 2
\begin{thm}[Result 3.2]
    Let $x \in \textbf{R}$.
    If $x ^ 2 - 2x + 2 \leq 0$, then $x ^ 3 \geq 8$.
\end{thm}
\begin{verbatim}
    Ex. Assume x is a real number. Assume x ^ 2 - 2 * x + 2 is less than or equal to 0.
    Then x ^ 3 is greater than or equal to 8.
\end{verbatim}
\begin{lstlisting}
    example (x : ℝ) (h57 : (((x ^ 2) - (2 * x)) + 2) ≤ 0) : (x ^ 3) ≥ 8 := sorry
\end{lstlisting}

\hrule
%Theorem 4
\begin{thm}[Exercise 3.1]
    Let $x \in \textbf{R}$.
    If $0 <  x < 1$, then $x^2 - 2x + 2 \neq 0$.
\end{thm}
\begin{verbatim}
    Ex. Assume x is a real number. Assume x is greater than 0 and x is less than 1.
    Then x ^ 2 - 2 * x + 2 is not equal to 0.
\end{verbatim}
\begin{lstlisting}
    example (x : ℝ) (h64 : x > 0) (h51 : x < 1) : (((x ^ 2) - (2 * x)) + 2) ≠ 0 := sorry
    example (x : ℝ) (h68 : x > 0) (h55 : x < 1) : (¬ (((x ^ 2) - (2 * x)) + 2) = 0) := sorry
\end{lstlisting}
In this case, GFLean produces two outputs which are syntactically same as Lean expressions.
This happens because the input produces two different parse trees.
Ideally, we would want a single parse tree.
\vspace{5mm}

\hrule
%Theorem 6
\begin{thm}[Exercise 3.3]
    Let $r \in \textbf{Q}^+.$
    If $\frac{r^2 + 1}{r} \leq 1$, then $\frac{r^2 + 2}{r} \leq 2$.
\end{thm}
\begin{verbatim}
    Ex. Assume r is a positive rational number. Assume (r ^ 2 + 1) / r is less than or
    equal to 1. Then (r ^ 2 + 2) / r is less than or equal to 2.
\end{verbatim}
\begin{lstlisting}
    example (r : ℚ) (h76 : pos r) (h63 : (((r ^ 2) + 1) / r) ≤ 1) : (((r ^ 2) + 2) / r) ≤ 2 := sorry
\end{lstlisting}

\hrule
%Theorem 7
\begin{thm}[Exercise 3.4]
    Let $x \in \textbf{R}$.
    If $x ^ 3 - 5 x - 1 \geq 0$, then $(x-1)(x-3) \geq -2.$
\end{thm}
\begin{verbatim}
    Ex. Assume x is a real number. Assume x ^ 3 - 5 * x - 1 is greater than or equal to 0.
    Then (x - 1) * (x - 3) is greater than or equal to -2.
\end{verbatim}
\begin{lstlisting}
    example (x : ℝ) (h70 : (((x ^ 3) - (5 * x)) - 1) ≥ 0) : ((x - 1) * (x - 3)) ≥ -2 := sorry
\end{lstlisting}

\hrule
% Theorem 9
\begin{thm}[Exercise 3.6]
    If $a, b$ and $c$ are odd integers such that $a + b + c = 0$, then $abc < 0$.
\end{thm}
\begin{verbatim}
    Ex. Assume a is an odd integer, b is an odd integer and c is an odd integer.
    Assume a + b + c is equal to 0. Then a * b * c is less than 0.
\end{verbatim}
\begin{lstlisting}
    example (a : ℤ) (h106 : odd a) (b : ℤ) (h85 : odd b) (c : ℤ) (h64 : odd c) (h51 : ((a + b) + c) = 0) : ((a * b) * c) < 0 := sorry
\end{lstlisting}

\hrule
% Theorem 10
\begin{thm}[Exercise 3.7]
    If $x, y$ and $z$ are three real numbers such that $x^2 + y^2 + z^{2} < xy + xz + yz$, then $x + y + z > 0$.
\end{thm}
\begin{verbatim}
    Ex. Assume x is a real number, y is a real number and z is a real number.
    Assume x ^ 2 + y ^ 2 + z ^ 2 is less than x * y + x * z + y * z. Then
    x + y + z is greater than 0.    
\end{verbatim}
\begin{lstlisting}
    example (x : ℝ) (y : ℝ) (z : ℝ) (h99 : (((x ^ 2) + (y ^ 2)) + (z ^ 2)) < (((x * y) + (x * z)) + (y * z))) : ((x + y) + z) > 0 := sorry
\end{lstlisting}

\hrule 
% Theorem 11
\begin{thm}[Result 3.4]
    If $n$ is an odd integer, then $3n + 7$ is an even integer.
\end{thm}
\begin{verbatim}
    Ex. Assume n is an odd integer. Then 3 * n + 7 is even.
\end{verbatim}
\begin{lstlisting}
    example (n : ℤ) (h40 : odd n) : even ((3 * n) + 7) := sorry
\end{lstlisting}

\hrule
%Theorem 12
\begin{thm}[Result 3.5]
    If $n$ is an even integer, then $-5n - 3$ is an odd integer.
\end{thm}
\begin{verbatim}
    Ex. Assume n is an even integer. Then -5 * n - 3 is odd.
\end{verbatim}
\begin{lstlisting}
    example (n : ℤ) (h41 : even n) : odd ((-5 * n) - 3) := sorry
\end{lstlisting}

\hrule
%Theorem 13
\begin{thm}[Result 3.6]
    If $n$ is an odd integer, then $4n^3 + 2n - 1$ is odd.
\end{thm}
\begin{verbatim}
    Ex. Assume n is an odd integer. Then 4 * n ^ 3 + 2 * n - 1 is odd.
\end{verbatim}
\begin{lstlisting}
    example (n : ℤ) (h57 : odd n) : odd ((4 * (n ^ 3)) + ((2 * n) - 1)) := sorry
\end{lstlisting}

\hrule
%Theorem 15
\begin{thm}[Result 3.8]
    If $n$ is an even integer, then $3n^5$ is an even integer.
\end{thm}
\begin{verbatim}
    Ex. Assume n is an even integer. Then 3 * n ^ 5 is even.
\end{verbatim}
\begin{lstlisting}
    example (n : ℤ) (h41 : even n) : even (3 * (n ^ 5)) := sorry
\end{lstlisting}

\hrule
%Theorem 16
\begin{thm}[Exercise 3.8]
    If $x$ is an odd integer, then $9x + 5$ is even.
\end{thm}
\begin{verbatim}
    Ex. Assume x is an odd integer. Then 9 * x + 5 is even.
\end{verbatim}
\begin{lstlisting}
    example (x : ℤ) (h40 : odd x) : even ((9 * x) + 5) := sorry
\end{lstlisting}

\hrule
%Theorem 17
\begin{thm}[Exercise 3.9]
    If $x$ is an even integer, then $5x - 3$ is an odd integer.
\end{thm}
\begin{verbatim}
    Ex. Assume x is an even integer. Then 5 * x - 3 is odd.
\end{verbatim}
\begin{lstlisting}
    example (x : ℤ) (h40 : even x) : odd ((5 * x) - 3) := sorry
\end{lstlisting}

\hrule
%Theorem 18
\begin{thm}[Exercise 3.10]
    If $a$ and $c$ are odd integers, then $ab + ac$ is even for every integer $b$.
\end{thm}
\begin{verbatim}
    Ex. Assume a is an odd integer and c is an odd integer. Then for every integer b,
    a * b + a * c is even.
\end{verbatim}
\begin{lstlisting}
    example (a : ℤ) (h78 : odd a) (c : ℤ) (h57 : odd c) :
    ∀ (b : ℤ), even ((a * b) + (a * c)) := sorry
\end{lstlisting}

\hrule
%Theorem 19
\begin{thm}[Exercise 3.11]
    Let $n \in \textbf{Z}$.
    If $1 - n ^2 > 0$, then $3 n - 2$ is an even integer.
\end{thm}
\begin{verbatim}
    Ex. Assume n is an integer. If 1 - n ^ 2 is greater than 0 then 3 * n - 2 is even.
\end{verbatim}
\begin{lstlisting}
    example (n : ℤ) : ((1 - (n ^ 2)) > 0 → even ((3 * n) - 2)) := sorry
\end{lstlisting}

\hrule
%Theorem 24
\begin{thm}[Result 3.10]
    Let $x \in \textbf{Z}$.
    If $5x - 7$ is even, then $x$ is odd.
\end{thm}
\begin{verbatim}
    Ex. Assume x is an integer. If 5 * x - 7 is even then x is odd.
\end{verbatim}
\begin{lstlisting}
    example (x : ℤ) : (even ((5 * x) - 7) → odd x) := sorry
\end{lstlisting}

\hrule
%Theorem 25
\begin{thm}[Result 3.11]
    Let $x \in \textbf{Z}$.
    Then $11 x - 7$ is even if and only if $x$ is odd.
\end{thm}
\begin{verbatim}
    Ex. Assume x is an integer. Then 11 * x - 7 is even iff x is odd.
\end{verbatim}
\begin{lstlisting}
    example (x : ℤ) : (even ((11 * x) - 7) ↔ odd x) := sorry
\end{lstlisting}

\hrule
%Theorem 26
\begin{thm}[Result 3.12]
    Let $x \in \textbf{Z}$.
    Then $x ^ 2$ is even if and only if $x$ is even.
\end{thm}
\begin{verbatim}
    Ex. Assume x is an integer. Then x ^ 2 is even iff x is even.
\end{verbatim}
\begin{lstlisting}
    example (x : ℤ) : (even (x ^ 2) ↔ even x) := sorry
\end{lstlisting}

\hrule
%Theorem 27
\begin{thm}[Lemma 3.13]
    Let $x \in \textbf{Z}$.
    If $5 x - 7$ is odd, then $x$ is even.
\end{thm}
\begin{verbatim}
    Ex. Assume x is an integer. If 5 * x - 7 is odd then x is even.
\end{verbatim}
\begin{lstlisting}
    example (x : ℤ) : (odd ((5 * x) - 7) → even x) := sorry
\end{lstlisting}

\hrule
%Theorem 28
\begin{thm}[Result 3.14]
    Let $x \in \textbf{Z}$.
    If $5x - 7$ is odd, then $9 x + 2$ is even.
\end{thm}
\begin{verbatim}
    Ex. Assume x is an integer. If 5 * x - 7 is odd then 9 * x + 2 is even.
\end{verbatim}
\begin{lstlisting}
    example (x : ℤ) : (odd ((5 * x) - 7) → even ((9 * x) + 2)) := sorry
\end{lstlisting}

\hrule
%Theorem 29
\begin{thm}[Exercise 3.16]
    Let $x \in \textbf{Z}$.
    If $7 x + 5$ is odd, then $x$ is even.
\end{thm}
\begin{verbatim}
    Ex. Assume x is an integer. If 7 * x + 5 is odd then x is even.
\end{verbatim}
\begin{lstlisting}
    example (x : ℤ) : (odd ((7 * x) + 5) → even x) := sorry
\end{lstlisting}

\hrule
%Theorem 30
\begin{thm}[Exercise 3.17]
    Let $n \in \textbf{Z}$.
    If $15n$ is even, then $9n$ is even.
\end{thm}
\begin{verbatim}
    Ex. Assume n is an integer. If 15 * n is even then 9 * n is even.
\end{verbatim}
\begin{lstlisting}
    example (n : ℤ) : (even (15 * n) → even (9 * n)) := sorry
\end{lstlisting}

\hrule
%Theorem 31
\begin{thm}[Exercise 3.18]
    Let $x \in \textbf{Z}$.
    Then $5x - 11$ is even if and only if $x$ is odd.
\end{thm}
\begin{verbatim}
    Ex. Assume x is an integer. Then 5 * x - 11 is even iff x is odd.
\end{verbatim}
\begin{lstlisting}
    example (x : ℤ) : (even ((5 * x) - 11) ↔ odd x) := sorry
\end{lstlisting}

\hrule
%Theorem 32
\begin{thm}[Exercise 3.19]
    Let $x \in \textbf{Z}$.
    If $7x + 4$ is even, then $3x - 11$ is odd.
\end{thm}
\begin{verbatim}
    Ex. Assume x is an integer. If 7 * x + 4 is even then 3 * x - 11 is odd.
\end{verbatim}
\begin{lstlisting}
    example (x : ℤ) : (even ((7 * x) + 4) → odd ((3 * x) - 11)) := sorry
\end{lstlisting}

\hrule
%Theorem 33
\begin{thm}[Exercise 3.20]
    Let $x \in \textbf{Z}$.
    Then $3x + 1$ is even if and only if $5x - 2$ is odd.
\end{thm}
\begin{verbatim}
    Ex. Assume x is an integer. Then 3 * x + 1 is even iff 5 * x - 2 is odd.
\end{verbatim}
\begin{lstlisting}
    example (x : ℤ) : (even ((3 * x) + 1) ↔ odd ((5 * x) - 2)) := sorry
\end{lstlisting}

\hrule
%Theorem 34
\begin{thm}[Exercise 3.21]
    Let $n \in \textbf{Z}$.
    Then $(n + 1)^2 - 1$ is even if and only if $n$ is odd.
\end{thm}
\begin{verbatim}
    Ex. Assume n is an integer. Then (n + 1) ^ 2 - 1 is even iff n is odd.
\end{verbatim}
\begin{lstlisting}
    example (n : ℤ) : (even (((n + 1) ^ 2) - 1) ↔ odd n) := sorry
\end{lstlisting}

\hrule
%Theorem 39
\begin{thm}[Result 3.15]
    If $n \in \textbf{Z}$, then $n^2 + 3n + 5$ is an odd integer.
\end{thm}
\begin{verbatim}
    Ex. Assume n is an integer. Then n ^ 2 + 3 * n + 5 is odd.
\end{verbatim}
\begin{lstlisting}
    example (n : ℤ) : odd (((n ^ 2) + (3 * n)) + 5) := sorry
\end{lstlisting}

\hrule
%Theorem 41
\begin{thm}[Theorem 3.17]
     Let $a$ and $b$ be integers.
     Then $ab$ is even if and only if $a$ is even or $b$ is even.
\end{thm}
\begin{verbatim}
    Ex. Assume a is an integer and b is an integer. Then a * b is even iff a is even or
    b is even.
\end{verbatim}
\begin{lstlisting}
    example (a : ℤ) (b : ℤ) : (even (a * b) ↔ (even a ∨ even b)) := sorry
\end{lstlisting}

\hrule
%Theorem 43
\begin{thm}[Exercise 3.26]
     If $n \in \textbf{Z}$, then $n^2 - 3n + 9$ is odd.
\end{thm}
\begin{verbatim}
    Ex. Assume n is an integer. Then n ^ 2 - 3 * n + 9 is odd.
\end{verbatim}
\begin{lstlisting}
    example (n : ℤ) : odd (((n ^ 2) - (3 * n)) + 9) := sorry
\end{lstlisting}

\hrule
%Theorem 44
\begin{thm}[Exercise 3.27]
    If $n \in \textbf{Z}$, then $n^3 - n$ is even.
\end{thm}
\begin{verbatim}
    Ex. Assume n is an integer. Then n ^ 3 - n is even.
\end{verbatim}
\begin{lstlisting}
    example (n : ℤ) : even ((n ^ 3) - n) := sorry
\end{lstlisting}

\hrule
%Theorem 45
\begin{thm}[Exercise 3.28]
    Let $x, y \in \textbf{Z}$.
    If $xy$ is odd, then $x$ and $y$ are odd.
\end{thm}
\begin{verbatim}
    Ex. Assume x is an integer and y is an integer. If x * y is odd then x is odd and
    y is odd.
\end{verbatim}
\begin{lstlisting}
    example (x : ℤ) (y : ℤ) : (odd (x * y) → (odd x ∧ odd y)) := sorry
\end{lstlisting}

\hrule
%Theorem 46
\begin{thm}[Exercise 3.29]
    Let $a, b \in \textbf{Z}$.
    If $ab$ is odd, then $a^2 + b^2$ is even.
\end{thm}
\begin{verbatim}
    Ex. Assume a is an integer and b is an integer. If a * b is odd then a ^ 2 + b ^ 2
    is even.
\end{verbatim}
\begin{lstlisting}
    example (a : ℤ) (b : ℤ) : (odd (a * b) → even ((a ^ 2) + (b ^ 2))) := sorry
\end{lstlisting}

\hrule
%Theorem 53
\begin{thm}[Exercise 3.36]
    Let $x, y \in \textbf{Z}$.
    If $3x + 4y$ and $4x + 5y$ are both even, then $x$ and $y$ are both even.
\end{thm}
\begin{verbatim}
    Ex. Assume x is an integer and y is an integer. If 3 * x + 4 * y is even and
    4 * x + 5 * y is even then x is even and y is even.
\end{verbatim}
\begin{lstlisting}
    example (x : ℤ) (y : ℤ) : ((even ((3 * x) + (4 * y)) ∧ even ((4 * x) + (5 * y))) → (even x ∧ even y)) := sorry
\end{lstlisting}

\hrule
%Theorem 54
\begin{thm}[Exercise 3.37]
    Let $x, y, z \in \textbf{Z}$.
    If exactly two of the three integers $x, y, z$ are even, then $3x + 5y + 7z$ is odd.
\end{thm}
\begin{verbatim}
    Ex. Assume x is an integer, y is an integer and z is an integer. Assume x is even,
    y is even and z is not even or x is even, y is not even and z is even or x is not
    even, y is even and z is even. Then 3 * x + 5 * y + 7 * z is odd.
\end{verbatim}
\begin{lstlisting}
    example (x : ℤ) (y : ℤ) (z : ℤ) (h158 : ((even x ∧ (even y ∧ (¬ even z))) ∨ ((even x ∧ ((¬ even y) ∧ even z)) ∨ ((¬ even x) ∧ (even y ∧ even z))))) : odd (((3 * x) + (5 * y)) + (7 * z)) := sorry
\end{lstlisting}

\hrule
%Theorem 57
\begin{thm}[Exercise 3.40]
    Let $a, b \in \textbf{Z}$.
    If $a$ is even or $b$ is even, then $ab$ is even.
\end{thm}
\begin{verbatim}
    Ex. Assume a is an integer and b is an integer. If a is even or b is even then
    a * b is even.
\end{verbatim}
\begin{lstlisting}
    example (a : ℤ) (b : ℤ) : ((even a ∨ even b) → even (a * b)) := sorry
\end{lstlisting}

\hrule
%Theorem 58
\begin{thm}[Example 3.19 (2)]
    If $n$ is an odd integer, then $3n - 5$ is an even integer.
\end{thm}
\begin{verbatim}
    Ex. Assume n is an odd integer. Then 3 * n - 5 is even.
\end{verbatim}
\begin{lstlisting}
    example (n : ℤ) (h40 : odd n) : even ((3 * n) - 5) := sorry
\end{lstlisting}

\hrule
%Theorem 59
\begin{thm}[Example 3.19 (4)]
    Let $n$ be an integer.
    If $3n-5$ is an odd integer, then $n$ is an even integer.
\end{thm}
\begin{verbatim}
    Ex. Assume n is an integer. If 3 * n - 5 is odd then n is even.
\end{verbatim}
\begin{lstlisting}
    example (n : ℤ) : (odd ((3 * n) - 5) → even n) := sorry
\end{lstlisting}

\hrule
%Theorem 61
\begin{thm}[Problem 3.21]
    If $m$ is an even integer and $n$ is an odd integer, then $3m + 5n$ is odd.
\end{thm}
\begin{verbatim}
    Ex. Assume m is an even integer and n is an odd integer. Then 3 * m + 5 * n is odd.
\end{verbatim}
\begin{lstlisting}
    example (m : ℤ) (h67 : even m) (n : ℤ) (h45 : odd n) : odd ((3 * m) + (5 * n)) := sorry
\end{lstlisting}

\hrule
%Theorem 63
\begin{thm}[Exercise 3.43]
    Let $n \in \textbf{Z}$.
    If $3n - 8$ is odd, then $n$ is odd.
\end{thm}
\begin{verbatim}
    Ex. Assume n is an integer. If 3 * n - 8 is odd then n is odd.
\end{verbatim}
\begin{lstlisting}
    example (n : ℤ) : (odd ((3 * n) - 8) → odd n) := sorry
\end{lstlisting}

\hrule
%Theorem 65
\begin{thm}[Exercise 3.45]
    Let $x, y \in \textbf{Z}$.
    If $x$ or $y$ is even, then $xy^2$ is even.
\end{thm}
\begin{verbatim}
    Ex. Assume x is an integer and y is an integer. If x is even or y is even then
    x * y ^ 2 is even.
\end{verbatim}
\begin{lstlisting}
    example (x : ℤ) (y : ℤ) : ((even x ∨ even y) → even (x * (y ^ 2))) := sorry
\end{lstlisting}

\hrule
%Theorem 66
\begin{thm}[Exercise 3.47]
    Let $x \in \textbf{Z}$.
    If $7x - 3$ is even, then $3x + 8$ is odd.
\end{thm}
\begin{verbatim}
    Ex. Assume x is an integer. If 7 * x - 3 is even then 3 * x + 8 is odd.
\end{verbatim}
\begin{lstlisting}
    example (x : ℤ) : (even ((7 * x) - 3) → odd ((3 * x) + 8)) := sorry
\end{lstlisting}

\hrule
%Theorem 67
\begin{thm}[Exercise 3.48]
    Let $n \in \textbf{Z}$.
    Then $(n-5)(n+7)(n+13)$ is odd if and only if $n$ is even.
\end{thm}
\begin{verbatim}
    Ex. Assume n is an integer. Then (n - 5) * (n + 7) * (n + 13) is odd iff n is even.
\end{verbatim}
\begin{lstlisting}
    example (n : ℤ) : (odd (((n - 5) * (n + 7)) * (n + 13)) ↔ even n) := sorry
\end{lstlisting}
\hrule

\section{A Formal Grammar of Simplified ForTheL} \label{grammar}
In this section, we give a formal grammar for Simplified ForTheL.
Although, we wrote a Grammatical Framework (GF) grammar for Simplified ForTheL, for comprehensibility, here we present it as a context-free grammar (CFG).
To present the Simplified ForTheL grammar as a CFG, we had to combine the abstract and concrete syntax in a way such that the readability is maintained.
As a result, the language defined by the following CFG is not exactly Simplified ForTheL, but a close approximation of it.
An exact CFG for Simplified ForTheL can be obtained by importing the \texttt{TextsEng.gf} file in the GF shell and typing the command \texttt{pg -printer=bnf} in the shell.

We use the BNF notation to present syntax.
Nonterminals are written in italic (e.g. \textit{variable}) and terminals in typewriter font (e.g. \texttt{integer}).
Grammar productions have the form:
\begin{align*}
    nonterm & \rightarrow  alt_1 \; | \; alt_2 \; | \; \dots \; | \; alt_n
\end{align*}
Let $t_1, t_2, t_3$ and $t_4$ be strings made up of terminals and non-terminals.
Then, the following conventions are adopted:
\begin{itemize}
    \item The symbol $\varepsilon$ denotes the empty string.
    \item The pattern $\; t_1 \; | \; t_2$ denotes a choice between $t_1$ and $t_2$.
    \item The pattern $t_1 \; [ \; t_2 \;] \; t_3$ denotes that $t_2$ is optional.
    \item The pattern $t_1 \; ( \; t_2 \; | \; t_3 \; ) \; t_4$ denotes a choice between $t_1 \; t_2 \; t_4$ and $t_1 \; t_3 \; t_4$.
\end{itemize}

We present the grammar in a bottom-up fashion. 
The following subsections, called Lexicon (\ref{lexicon}), Notions (\ref{notions}), Terms (\ref{terms}), Predicates (\ref{predicates}), Statements (\ref{statements}), and Texts (\ref{texts}), correspond to the Abstract Syntax file names found in the GitHub repository of the project \cite{gflean}.
With respect to the grammar given, GFLean works on a \textit{text} and produces its formalisation.
\subsection{Lexicon} \label{lexicon}

\begin{center}
\begin{align*}
    variable & \rightarrow \; \texttt{a} \; \; | \; \; \texttt{b} \; \; | \; \; \texttt{c} \; \; | \; \; \texttt{k} \; \; | \; \; \texttt{m} \; \; | \; \; \texttt{n} \; \; | \; \; \texttt{r} \; \; | \; \; \texttt{x} \; \; | \; \; \texttt{y} \; \; | \; \; \texttt{z} \\
    rawNoun0 & \rightarrow \texttt{real} \; (\; \texttt{number} \; | \; \texttt{numbers} \;) \\
        & \; \; | \; \; ( \; \texttt{integer} \; | \; \texttt{integers} \; ) \\
        & \; \; | \; \; \texttt{rational} \; (\; \texttt{number} \; | \; \texttt{numbers} \;) \\
    rawAdjective1 & \rightarrow \texttt{less than} \\
        & \; \; | \; \; \texttt{less than or equal to} \\
        & \; \; | \; \; \texttt{greater than} \\
        & \; \; | \; \; \texttt{greater than or equal to} \\
        & \; \; | \; \; \texttt{not equal to} \\
        & \; \; | \; \; \texttt{equal to} \\
    rawAdjective0 & \rightarrow \; \texttt{positive} \; | \; \texttt{odd} \; | \; \texttt{even} \; | \; \texttt{nonnegative} \; | \; \texttt{negative} \\
    rawNoun2 & \rightarrow \; \texttt{+} \; | \; \texttt{-} \; | \; \texttt{*} \; | \; \texttt{/}  \; | \; \texttt{\textasciicircum}
\end{align*}
\end{center}

\subsection{Notions} \label{notions}
\begin{center}
\begin{align*}
    primSimpleAdjective & \rightarrow rawAdjective0 \\
    primClassNoun & \rightarrow rawNoun0 \; names \\
    names & \rightarrow variable \\
    variable & \rightarrow \texttt{(x} \; Int \texttt{)} \\
        & \; \; | \; \; \varepsilon \\
    leftAttribute & \rightarrow primSimpleAdjective \\
    rightAttribute & \rightarrow isPredicate \\
        & \; \; | \; \; \texttt{that} \; doesPredicate \\
        & \; \; | \; \; \texttt{such that} \; statement \\
    notion & \rightarrow primClassNoun \\
        & \; \; | \; \; primClassNoun \; rightAttribute \\
        & \; \; | \; \; leftAttribute \; primClassNoun \\
        & \; \; | \; \; leftAttribute \; primClassNoun \; rightAttribute \\
    Int & \rightarrow \; \dots \; |  \; \texttt{-1}  \; |  \; \texttt{0}  \; |  \; \texttt{1}  \; |  \; \dots
\end{align*}
\end{center}
    
\subsection{Terms} \label{terms}
\begin{center}
\begin{align*}
    primDefiniteNoun & \rightarrow term \; rawNoun2 \; term \\
    term & \rightarrow quantifiedNotion \\
        & \; \; | \; \; definiteTerm \\
    quantifiedNotion & \rightarrow \texttt{every} \; term \\
        & \; \; | \; \; \texttt{some} \; term \\
        & \; \; | \; \; \texttt{no} \; term \\
    definiteTerm & \rightarrow primDefiniteNoun \\
        & \; \; | \; \; variable \\
        & \; \; | \; \; Int 
\end{align*}
\end{center}

\subsection{Predicates} \label{predicates}
\begin{center}
\begin{align*}
    polarity & \rightarrow \epsilon \; | \; \texttt{not} \\
    primAdjective & \rightarrow rawAdjective0 \\
        & \; \; | \; \; rawAdjective1 \; term \\
    doesPredicate & \rightarrow ( \; \texttt{is} \; | \; \texttt{are} \; ) \; isPredicate \\
        & \; \; | \; \; ( \; \texttt{is} \; | \; \texttt{are} \; ) \; is\_aPredicate \\
    isPredicate & \rightarrow polarity \; primAdjective \\
    is\_aPredicate & \rightarrow polarity \; ( \; \texttt{a} \; | \; \texttt{an} \; | \; \epsilon \; ) \; notion \\
        & \; \; | \; \; polarity \; definiteTerm
\end{align*}
\end{center}

\subsection{Statements} \label{statements}
\begin{center}
\begin{align*}
    statement & \rightarrow statement \; ( \; \texttt{and} \; | \; \texttt{,} \; ) \; statement \\
            & \; \; | \; \; statement \; \texttt{or} \; statement \\
            & \; \; | \; \; \texttt{if} \; statement \; \texttt{then} \; statement \\
            & \; \; | \; \; \texttt{it's not that} \; statement \\
            & \; \; | \; \; \texttt{for} \; quantifiedNotion \; \texttt{,} \; statement \\
            & \; \; | \; \; term \; doesPredicate \\
            & \; \; | \; \; ( \; \texttt{there exist} \; | \; \texttt{there exists a} \; | \;\texttt{there exists an} \; ) \; notion \\
            & \; \; | \; \; (\; \texttt{there exists no} \; | \; \texttt{there exist no} \; ) \; notion
\end{align*}
\end{center}

\subsection{Texts} \label{texts}
\begin{center}
\begin{align*}
    text & \rightarrow example \\
    example & \rightarrow \texttt{ex.} \; Lassumption \; [\texttt{then}] \; statement \; \texttt{.} \\
    Lassumption & \rightarrow  \texttt{assume} \; assumption \; Lassumption \;  |  \; \varepsilon  \\
    assumption & \rightarrow statement \; \texttt{.}
\end{align*}
\end{center}

\end{document}